\documentclass[runningheads]{llncs}
\usepackage{graphicx}
\usepackage{epsfig}
\usepackage{amsmath}
\usepackage{cases}

\usepackage{amsfonts}      
\usepackage{xcolor}
\usepackage[utf8]{inputenc}
\newcommand{\B}{\boldsymbol}
\usepackage{cleveref}
\usepackage{subcaption}
\begin{document}
\title{On the relation between\\Loss Functions and T-Norms\thanks{This project has received funding from the European Union's Horizon 2020 research and innovation program under grant agreement No 825619.}
\thanks{To appear in 29th International Conference on Inductive Logic Programming, Plovdiv, Bulgaria on 3-5 Sep 2019.  \textsuperscript{\textcopyright} ILP2019}}
\author{Francesco Giannini\inst{1} \and
Giuseppe Marra\inst{1,2} \and
Michelangelo Diligenti\inst{1} \and \\
Marco Maggini\inst{1} \and
Marco Gori\inst{1}}
\authorrunning{F. Giannini et al.}
\institute{Department of Information Engineering and Mathematical Sciences, \\ University of Siena, ITALY \\ \email{\{fgiannini,diligmic,maggini,marco\}@diism.unisi.it}\\	
	 \and Department of Information Engineering, \\ University of Florence, ITALY \\
\email{g.marra@unifi.it}
}

\maketitle          
\begin{abstract}
Deep learning has been shown to achieve impressive results in several domains like computer vision and natural language processing.
A key element of this success has been the development of new loss functions, like the popular cross-entropy loss, which has been shown to provide faster convergence and to reduce the vanishing gradient problem in very deep structures.
While the cross-entropy loss is usually justified from a probabilistic perspective, this paper shows an alternative and more direct interpretation of this loss in terms of t-norms and their associated generator functions, and derives a general relation between loss functions and t-norms.
In particular, the presented work shows intriguing results leading to the development of a novel class of loss functions. These losses can be exploited in any supervised learning task and which could lead to faster convergence rates that the commonly employed cross-entropy loss.

\end{abstract}

\section{Introduction}
A careful choice of the loss function has been pivotal into the success of deep learning.
In particular, the {\bf cross-entropy loss}, or log loss, measures the performance of a classifier and increases when the predicted probability of an assignment diverges from the actual label~\cite{goodfellow2016deep}. In supervised learning, the cross-entropy loss has a clear interpretation as it attempts at minimizing the distribution of the predicted and given pattern labels. From a practical standpoint, the main advantage of this loss is to limit the vanishing gradient issue for networks with sigmoidal or softmax output activations.

Recent advancements in Statistical Relational Learning (SRL)~\cite{koller2007introduction} allow to inject prior knowledge, often expressed using a logic formalism, into a learner.
One of the most popular lines of research in this community attempts at defining frameworks for performing logic inference in the presence of uncertainty. For example, Markov Logic Networks~\cite{richardson2006markov} and Probabilistic Soft Logic~\cite{bach2017hinge} integrate First Order Logic (FOL) and graphical models. More recently, many attempts have been focusing on integrating reasoning with uncertainty with deep learning~\cite{xu2017semantic}. A common solution, followed by approaches like Semantic Based Regularization~\cite{diligenti2017semantic} and Logic Tensor Networks~\cite{donadello2017logic}, relies on using deep networks to approximate the FOL predicates, and the overall architecture is optimized end-to-end by relaxing the FOL into a differentiable form, which translates into a set of constraints.
For the sake of overall consistency, one question that can naturally arise in this context is how  the fitting of the supervised examples can be expressed using logic formalism. Following this starting point, this paper follows an  orthogonal approach for the definition of a loss function, by studying the relation between the translation of the prior knowledge using t-norms and the resulting loss function. In particular, the notion of t-norm \emph{generator} plays a fundamental role in the behavior of the corresponding loss.
Remarkably, the cross-entropy loss can be naturally derived within this framework. However, the presented theoretical results suggest that there is a larger class of loss functions that correspond to the different possible translations of logic using t-norms, and some loss functions are potentially more effective than the cross-entropy to limit the vanishing gradient issue, therefore proving a faster convergence rate.

The paper is organized as follows: Section~\ref{sec:aggreg} presents the basic concepts about t-norms, generators and aggregator functions. Section \ref{sec:learning_from_constraints} introduces the learning frameworks used to represent supervised learning in terms of logic rules, while Section~\ref{sec:experiments} presents the experimental results and, finally, Section~\ref{sec:conclusions} draws some conclusions.

\section{Fuzzy Aggregation Functions}
\label{sec:aggreg}
The aggregation takes place on a set of values typically representing preferences or satisfaction degrees restricted to the unit interval $[0,1]$ to be aggregated. There are several ways to aggregate them into a single value expressing an overall combined score, according to what is expected from such mappings.
The purpose of aggregation functions is to combine inputs that are typically interpreted as degrees of membership in fuzzy sets, degrees of preference or strength of evidence. Aggregation functions have been studied by several authors in the literature \cite{beliakov2007aggregation,calvo2002aggregation}, and they are successfully used in many practical applications, for instance see \cite{grabisch2011aggregation,torra2007modeling}.
Please note that the fuzzy aggregation functions that will be covered in this section can be directly applied to the output of a multi-task classifier, when implemented via a neural network with sigmoidal or softmax output units.

\subsubsection{Basic Definitions.}
Aggregation functions are defined for inputs of any cardinality, however for simplicity the main definitions are provided only for the binary case. A (binary) aggregation function is a non-decreasing function $A:[0,1]^2\rightarrow[0,1]$, such that: $A(0,0)=0$, $A(1,1)=1$.
An aggregation function $A$ can be categorized according to the pointwise order in Equation~\ref{eq:ordagg} as: \emph{conjunctive} when $A\leq\min$, \emph{disjunctive} when
$\max\leq A$, \emph{averaging} (a \emph{mean}) when $\min< A<\max$ and \emph{hybrid} otherwise; where $\min$ and $\max$ are the aggregation functions for the $minimum$ and $maximum$ respectively. 
\begin{equation}\label{eq:ordagg}
A_1\leq A_2\quad\mbox{iff}\quad A_1(x,y)\leq A_2(x,y), \mbox{ for all }x,y\in[0,1] \ .
\end{equation}

Conjunctive and disjunctive type functions combine values as if they were related by a logical AND and OR operations, respectively. On the other hand,
averaging type functions have the property that low values can be compensated by high values. Mean computation is the most common way to combine the inputs, since it assumed the total score cannot be above or below any of the inputs, but it depends on all the inputs.

\subsection{Archimedean T-Norms}
\label{sec:gent-norm}
Despite averaging functions have nice properties to aggregate fuzzy values, they are not suitable to represent neither a conjunction nor a disjunction, because they do not generalize their boolean counterpart. This is a reason why, we focus on t-norms and t-conorms~ \cite{klement2004triangular1,klement2013triangular}, that are \emph{associative, commutative} aggregation functions with 1 and 0 as \emph{neutral element}, respectively. 
Table \ref{tab:tnorms} reports G\"{o}del, Lukasiewicz and Product t-norms, which are referred as the fundamental t-norms because all the continuous t-norms can be obtained as ordinal sums of the two fundamental t-norms~\cite{jenei2002note}. 
A simple example of a t-norm that is not continuous is given by the Drastic t-norm $T_{D}$, that is always returning a zero value, except for $T_{D}(1,1) = 1$.
\emph{Archimedean} t-norms \cite{klement2004triangular3} are a class of t-norms that can be constructed by means of unary monotone functions, called \emph{generators}.
\begin{table}[t]
    \centering
    \begin{tabular}{|c|c|c|}
    \hline
    {\bf G\"{o}del} & {\bf Lukasiewicz} & {\bf Product} \\
    \hline
    \hline 
      $T_M(x,y)=\min\{x,y\}$   & $T_{L}(x,y) =\max\{0,x+y-1\}$ &  $T_{\Pi}(x,y) =x\cdot y$ \\
      \hline
    \end{tabular}
    \label{tab:my_label}
    \caption{Fundamental t-norms.}\label{tab:tnorms}
\end{table}

\begin{definition}
	A t-norm $T$ is said to be \emph{Archimedean} if for every $x\in(0,1)$, $T(x,x)<x$. In addition, $T$ is said \emph{strict} if for all $x\in(0,1)$, $0<T(x,x)<x$ otherwise is said \emph{nilpotent}.
\end{definition}
For instance, the Lukasiewicz t-norm $T_{L}$ is nilpotent, the Product t-norm $T_{\Pi}$ is strict, while the G\"{o}del one $T_M$ is not archimedean, indeed $T_M(x,x)=x$, for all $x\in[0,1]$. The Lukasiewicz and Product t-norms are enough to represent the whole classes of nilpotent and strict Archimedean t-norms~\cite{klement2013triangular}.

A fundamental result for the construction of t-norms by \emph{additive} generators is based on the following theorem~\cite{klement2004triangular2}:
\begin{theorem}
	Let $g:[0,1]\to[0,+\infty]$ be a strictly decreasing function with $g(1)=0$  and $g(x)+g(y)\in Range(g)\cup[g(0^+),+\infty]$ for all $x, y$ in $[0, 1]$, and $g^{(-1)}$ its pseudo-inverse. Then the function $T:[0,1]\to[0,1]$ defined as
\begin{equation}\label{eq:t-normgen}
	T(x,y)=g^{-1}\left(\min\{g(0^+),g(x)+g(y)\}\right) \ .
\end{equation}
	is a t-norm and $g$ is said an \emph{additive generator} for $T$.
\end{theorem}
Any t-norm $T$ with an additive generator $g$ is Archimedean, if $g$ is continuous then $T$ is continuous, $T$ is strict if and only if $g(0)=+\infty$, otherwise it is nilpotent. 

\begin{example}
	If we take $g(x)=1-x$, then also $g^{-1}(y)=1-y$ and we get $T_{L}$:
	\[
	T(x,y)=1-\min\{1,1-x+1-y\}=\max\{0,x+y-1\} \ .
	\]
\end{example}	
\begin{example}	
	\label{ex:genlog}
		Taking $g(x)=-\log(x)$, we have $g^{-1}(y)=e^{-y}$ and we get $T_{\Pi}$:
		\[
		T(x,y)=e^{-(\min\{+\infty,-\log(x)-\log(y)\})}= x\cdot y \ .
		\]
\end{example}
Eq. (\ref{eq:t-normgen}) allows to derive the other fuzzy connectives as function of the generator:
\begin{equation}
\label{eq:genres}
\begin{aligned}
\mbox{residuum}: & & x\Rightarrow y = g^{-1}\left(\max\{0,g(y)-g(x)\}\right) \\ 
\mbox{bi-residuum}: & & x\Leftrightarrow y = g^{-1}\left(|g(x)-g(y)|\right) 
\end{aligned}
\end{equation}

If $g$ is expressed as a parametric function, it is possible to define families of t-norms, which can be constructed by the generator obtained when setting the parameters to specific values. Several parametric families of t-norms have been introduced~\cite{beliakov2007aggregation}. The experimental section of this paper employs the family of Schweizer–Sklar and Frank t-norms, depending on a parameter  $\lambda\in(-\infty,+\infty)$ and $\lambda\in[0,+\infty]$ respectively, and whose generators are defined as:
\begin{equation}\label{eq:SS}
g_{\lambda}^{SS}(x)=
\begin{cases}
-\log(x) & \mbox{if }\lambda=0\\
\frac{1-x^{\lambda}}{\lambda} & \mbox{otherwise}
\end{cases}\;\;
\mbox{and}\;\;
g_{\lambda}^{F}(x)=
\begin{cases}
-\log(x) & \mbox{if }\lambda=1\\
1-x & \mbox{if }\lambda=+\infty\\
\log\left(\frac{\lambda-1}{\lambda^x-1}\right) & \mbox{otherwise}
\end{cases}
\end{equation}

\section{From Formulas to Loss Functions}
\label{sec:learning_from_constraints}
A learning process can be thought of as a constraint satisfaction problem, where the constraints represent the knowledge about the functions to be learned. In particular, multi-task learning can be expressed via a set of constraints expressing the fitting of the supervised examples, plus any additional abstract knowledge.



Let us consider a set of unknown task functions ${\bf P}=\{p_1,\ldots,p_J\}$ defined on $\mathbb{R}^n$, all collected in the vector $\B p=(p_1,\ldots,p_J)$ and a set of known functions or predicates $\B S$.
Given the set $\mathcal{X}\subseteq\mathbb{R}^n$ of available data, a learning problem can be generally formulated as
$\min_{\B p}\mathcal{L}(\mathcal{X},\B S,\B p)$
where $\mathcal{L}$ is a positive-valued functional denoting a certain loss function. 
Each predicate is approximated by a neural network providing an output value in $[0,1]$. 
The available knowledge about the task functions consists in a set of FOL formulas $KB=\{\varphi_1,\ldots,\varphi_H\}$
and the learning process aims at finding a good approximation of each unknown element, so that the estimated values will satisfy the formulas for the input samples.
Since any formula is true if it evaluates to 1, in order to satisfy the constraints we may minimize the following loss function:
\begin{equation}\label{eq:loss}
\mathcal{L}(\mathcal{X},\B S, \B p) = \sum_{h=1}^H \lambda_h L\big(f_h(\mathcal{X},\B S, \B p)\big)
\end{equation}
where any $\lambda_h$ is the weight for the $h$-th logical constraint, which can be selected via cross-validation or jointly learned \cite{kolb2018learning,yang2017differentiable}, $f_h$ is the truth-function corresponding to the formula $\varphi_h$ according to a certain t-norm fuzzy logic and $L$ is a decreasing function denoting the penalty associated to the distance from satisfaction of formulas, so that $L(1)=0$. In the following, we will study different forms for the $L$ cost function and how it depends on the choice of the t-norm generator.
In particular, a \emph{t-norm fuzzy logic} generalizes Boolean logic to variables assuming values in $[0,1]$ and is defined by its t-norm modeling the logical AND \cite{hajek2013metamathematics}. The connectives can be treated using the fuzzy generalization of first--order logic that was first proposed by Novak~\cite{novak2012mathematical}. The \emph{universal} and \emph{existential quantifiers} occurring in the formulas in $KB$ allows the aggregation of different evaluations (groundings) of the formulas on the available data. For instance, given a formula $\varphi(x_i)$ depending on a certain variable $x_i\in\mathcal{X}_i$, where $\mathcal{X}_i$ denotes the available samples for the $i$-th argument of one of the involved predicates in $\varphi$, we may convert the quantifiers as the minimum and maximum operations that are common to any t-norm fuzzy logic:
\begin{equation*}
\label{eq:FOL}
\begin{aligned}
\forall x_i\, \varphi(x_i) & \Longrightarrow & f_{\varphi}(X_i,\B S,\B p) =   \displaystyle\min_{x_i\in\mathcal{X}_i} f_{\varphi}(x_i,\B S,\B p)  \\ 
\exists x_i\, \varphi(x_i) & \Longrightarrow & f_{\varphi}(X_i,\B S, \B p) = \displaystyle\max_{x_i\in\mathcal{X}_i} f_{\varphi}(x_i,\B S,\B p) 
\end{aligned}
\end{equation*}

\subsection{Loss Functions by T-Norms Generators}
\label{sec:Lgen}
A quantifier can be seen as a way to aggregate all the possible groundings of a predicate variable that, in turn, are $[0,1]$-values.
Different aggregation functions have also been considered, for example in~\cite{donadello2017logic}, the authors consider a mean operator to convert the universal quantifier. However this has the drawback that also the existential quantifier has the same semantics conversion and then it is determined by the authors via Skolemization. 
Even if this choice may yield some learning benefits, it has no direct justification inside a logic theory. Moreover it does not suggest how to map the functional translation of the formula into a constraint. In the following, we investigate the mapping of formulas into constraints by means of generated t-norm fuzzy logics, and we exploited the same additive generator of the t-norm to map the formula into the functional constraints to be minimized, i.e. $L=g$.

Given a certain formula $\varphi(x)$ depending on a variable $x$ that ranges in the set $\mathcal{X}$ and its corresponding functional representation $f_{\varphi}(x,\B p)$ evaluated on each $x\in\mathcal{X}$, the conversion of universal and existential quantifiers should have semantics equivalent to the AND and OR of the evaluation of the formula over the groundings, respectively. This can be realized by directly applying the t-norm or t-conorms over the groundings. For instance, for the universal quantifier:
\begin{equation}\label{eq:constbygen}
	\forall x\,\varphi(x) \equiv \displaystyle\bigwedge_x \varphi(x) \quad\Longrightarrow\quad g^{-1}\left(\min\left\{g(0^+),\sum_{x\in\mathcal{X}}g\big(f_{\varphi}(x,\B S, \B p)\big)\right\}\right) \ ,
\end{equation}
where $g$ is an additive generator of the t-norm $T$ corresponding to the universal quantifier.
Since any generator function is decreasing, in order to maximize the satisfaction of $\forall x\,\varphi(x)$ we can minimize $g$ applied to Equation \ref{eq:constbygen}, namely:
\begin{eqnarray}
\min\{g(0^+),\displaystyle\sum_{x\in\mathcal{X}}g(f_{\varphi}(x,\B S, \B p))\} & \qquad\qquad & \mbox{if $T$ is nilpotent} \label{eq:nilgen}\\
\nonumber\\
\displaystyle\sum_{x\in\mathcal{X}}g(f_{\varphi}(x,\B S, \B p)) & \qquad\qquad & \mbox{if $T$ is strict} \label{eq:strictgen}
\end{eqnarray}

As a consequence, with respect to the convexity of the expressions in Equations \ref{eq:nilgen}-\ref{eq:strictgen},  we get the following result, that is an immediate consequence of how the convexity is preserved by function composition.
	\begin{proposition}
If $g$ is a linear function and $f_\varphi$ is concave, Equation \ref{eq:nilgen} is convex. 
If $g$ is a convex function and $f_\varphi$ is linear, Equation \ref{eq:strictgen} is convex. 
	\end{proposition}
	
	\begin{example}
		If $g(x)=1-x$ (Lukasiewicz t-norm) from Equation \ref{eq:nilgen} we get:
		\[
		\min(1,\sum_{x\in\mathcal{X}}(1-(f_\varphi(x,\B S, \B p))) \ .
		\]
		Hence, in case $f_\varphi$ is concave (see \cite{giannini2018convex} for a characterization of the concave fragment of Lukasiewicz logic), this function is convex.

\noindent		If $g=-\log$ (Product t-norm) from Equation \ref{eq:strictgen} we get the cross-entropy:
		\[
		-\sum_{x\in\mathcal{X}}\log(f_\varphi(x,\B S, \B p)) \ .
		\]
		\end{example}

As we already pointed out in Section \ref{sec:aggreg}, if $g$ is an additive generator for a t-norm $T$, then the residual implication and the biresidum with respect to $T$ are given by Equation \ref{eq:genres}.
In particular, if $p_1,p_2$ are two unary predicates functions sharing the same input domain $\mathcal{X}$, and $\B S=\emptyset$ the following formulas yield the following penalty terms:
\begin{eqnarray*}\label{eq:convFOLgen}
\forall x\,p_1(x) & \Longrightarrow & \displaystyle \min\left\{g(0^+),\sum_{x \in \mathcal{X}} g(p_1(x))\right\} \nonumber\\
\forall x\,p_1(x)\Rightarrow p_2(x) & \Longrightarrow & \displaystyle \min\left\{g(0^+),\sum_{x \in \mathcal{X}} \max(0, g(p_2(x))-g(p_1(x))\right\} \\
\forall x\,p_1(x)\Leftrightarrow p_2(x) & \Longrightarrow & \displaystyle \min\left\{g(0^+),\sum_{x \in \mathcal{X}} |g(p_1(x))-g(p_2(x))|\right\} \ . \nonumber
\end{eqnarray*}

\subsection{Redefinition of supervised Learning with Logic}
\label{sec:supervised_learning}
In this section, we study the case of supervised learning w.r.t. the choice of a certain additive generator.
Let us consider a multi-task classification problem with predicates $p_j, j=1,\dots,J$ defined over the same input domain  with a supervised training set $\mathcal{T}=\{(x_i,y_i)\}$ where each $y_i\in \{1,2,\ldots,J\}$ is the output class for the pattern $x_i$ and $\mathcal{X}$ is the overall set of supervised patterns. Finally, the known predicate $S_j$ is defined for each predicate such that $S_j(x_i)=1$ iff $y_i=j$, and we indicate as $\mathcal{X}_j = \{x_i \in \mathcal{X} : S_j(x_i)=1\}$ the set of positive examples for the $j$-th predicate. Then, we can enforce the supervision constraints for $p_j$ as:

\begin{equation*}
    \forall x\,S_j(x)\Leftrightarrow  p_j(x)\quad \Longrightarrow\quad \mathcal{L} (\mathcal{X}, \B S, p_j) = \sum_{x\in\mathcal{X}} |g(S_j(x)) - g(p_j(x))|
\end{equation*}

In the special case of the predicates implemented by neural networks and exclusive multi-task classification, where each pattern should be assigned to one and only one class, the exclusivity can be enforced using a softmax output activation.
Typically, in this scenario, only the positive supervisions are explicitly listed, and since it holds that $g(S_j(x))=0, \forall x \in \mathcal{X}_j$, yields:
\begin{equation}
\mathcal{L}^+ (\mathcal{X},\B S, p_j) = \sum_{x\in\mathcal{X}_j} g(p_j(x)),
\label{eq:pos_only}
\end{equation} 
For instance, in the case of Lukasiewicz and Product logic, we have, respectively: 
\begin{equation*}
    \mathcal{L}_{L}^+(\mathcal{X}_j,p_j) = \sum_{x\in\mathcal{X}_j} \left(1-p_j(x)\right), \quad \mathcal{L}_{\Pi}^+(\mathcal{X}_j,p_j) = -\sum_{x\in\mathcal{X}_j} \log\left(p_j(x)\right)
\end{equation*}
corresponding to the $L_1$ and cross entropy losses, respectively.

\section{Experimental Results}
\label{sec:experiments}
\begin{figure}[th]
\centering
\begin{subfigure}[t]{.49\textwidth}
  \centering
 \includegraphics[width=\linewidth]{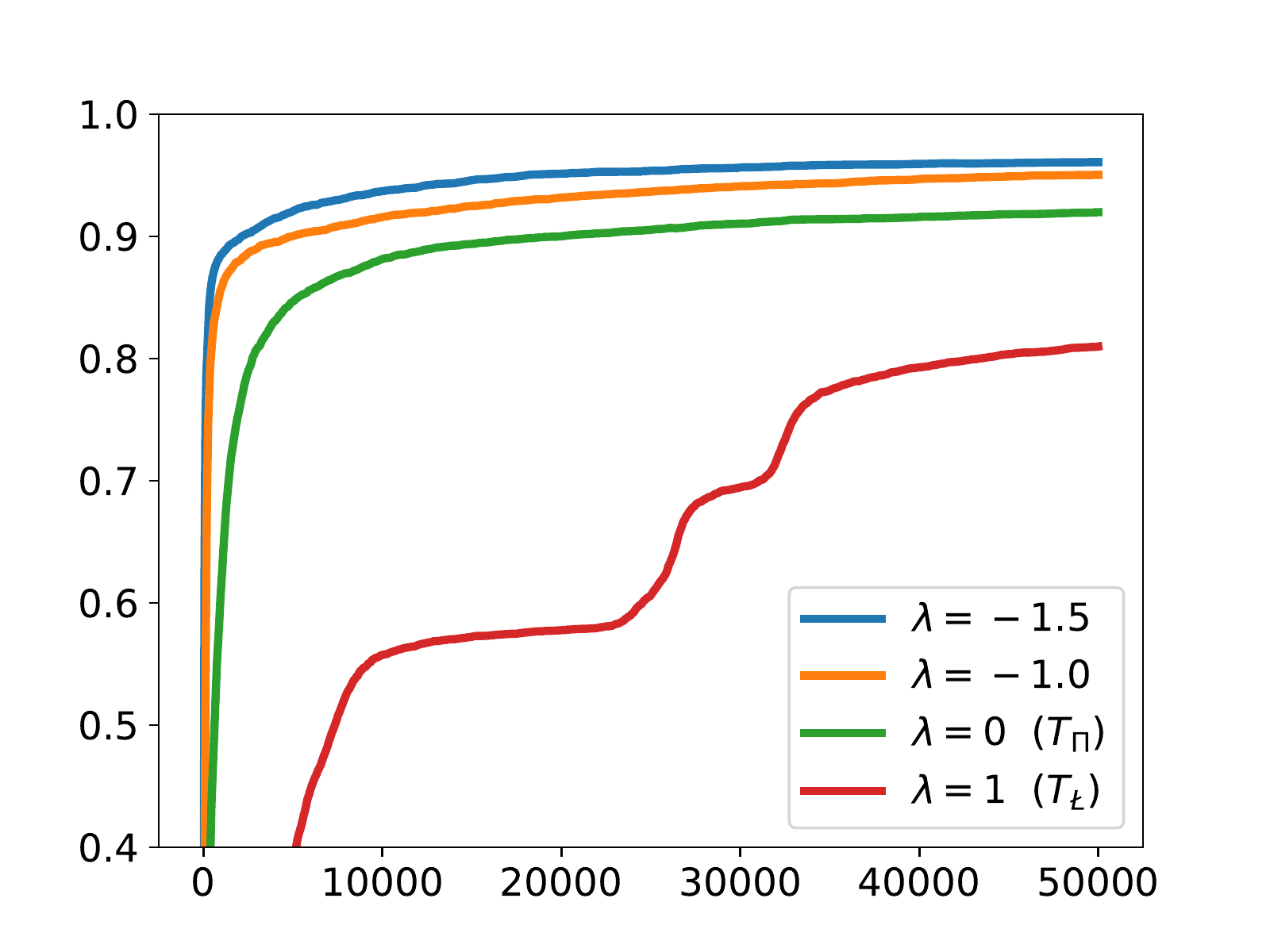}
 \caption{The Schweizer–Sklar t-norms}
  \nonumber
\end{subfigure}
\hfill
\begin{subfigure}[t]{.49\textwidth}
  \centering
 \includegraphics[width=\linewidth]{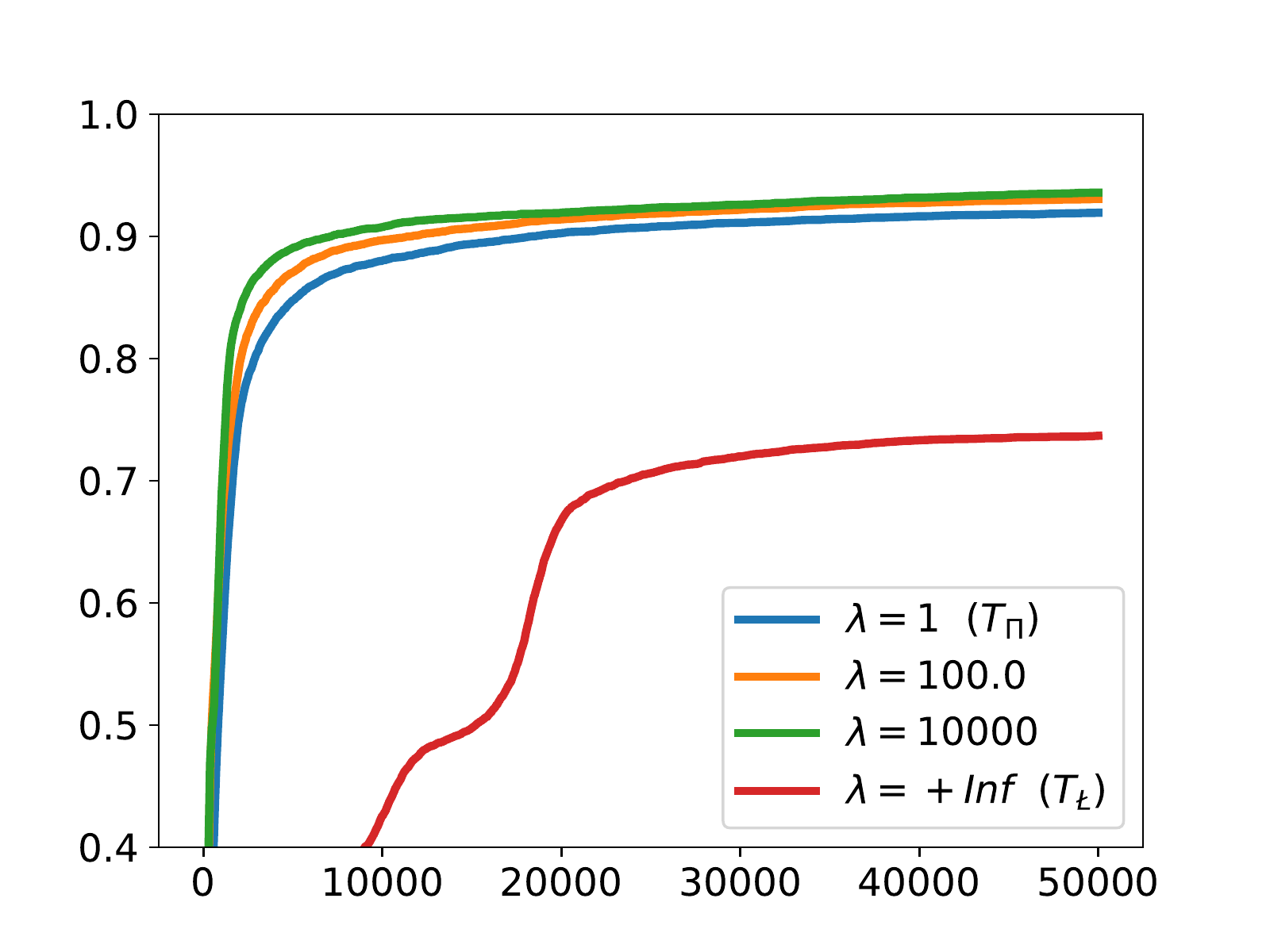}
 \caption{The Frank t-norms}
  \nonumber
\end{subfigure}
 \caption{Convergence speed of multiple generated loss functions on the MNIST classification task for different values of the parameter $\lambda$ of equation~\ref{eq:SS}. The well-known cross-entropy loss is equivalent to the loss obtained by the $T_{\Pi}$ generator.}
\label{fig:exp}
\end{figure}

The proposed framework allows to recover well-known loss functions by expressing the fitting of the supervision using logic and then carefully selecting the t-norm used to translate the resulting formulas. However, a main strength of the proposed theory is that it becomes possible to derive new principled losses starting from any family of parametric t-norms.
Driven by the huge impact that cross-entropy gained w.r.t. to classical loss functions in improving convergence speed and generalization capabilities, we designed a set of experiments to investigate how the choice of a t-norm can lead to a loss function with better performances than the cross-entropy loss. 
The Schweizer–Sklar and the Frank parametric t-norms defined in Section~\ref{sec:gent-norm} have been selected for this experimental evaluation, given the large spectrum of t-norms that can be generated by varying their $\lambda$ parameter. The well known MNIST dataset is used as benchmark for all the presented experiments. In order to have a fair comparison, the same neural network architecture is used during all the runs: a 1-hidden layer neural network with 50 hidden ReLU units and 10 softmax output units. The softmax activation function allows to express only positive supervisions, like commonly done in mutually exclusive classification using the cross-entropy loss. Optimization is carried on using Vanilla gradient descent with a fixed learning rate of $0.01$. 

Results are shown in Figure~\ref{fig:exp}, that reports the accuracy on the test set of a neural network trained on the MNIST dataset. 
Specific choices of the parameter $\lambda$ recover classical loss functions, like the cross-entropy loss, which is equivalent to the loss obtained using $T_{\Pi}$. The results confirm that the cross-entropy loss converges faster than the $L_1$ obtained when using $T_{L}$. However, there is a wide range of possible choices for the parameter $\lambda$ that brings an even faster convergence and better generalization than the widely adopted used cross-entropy.

\section{Conclusions}
\label{sec:conclusions}
This paper presents a framework to embed prior knowledge expressed as logic statements into a learning task, showning how the choice of the t-norm used to convert the logic into a differentiable form defines the resulting loss function used during learning.
When restricting the attention to supervised learning, the framework recovers popular loss functions like the cross-entropy loss, and allows to define new loss functions corresponding to the choice of the parameters of t-norm parametric forms. The experimental results show that some newly defined losses provide a faster convergence rate that the commonly used cross-entropy loss.
Future work will focus on testing the loss functions in more structured learning tasks, like the one commonly addressed with Logic Tensor Networks and Semantic based Regularization. The parametric form of the loss functions allows to define joint learning tasks, where the loss parameters are co-optimized during learning, for example using maximum likelihood estimators.

\bibliographystyle{splncs04}
\bibliography{references}

\end{document}